# A review of machine learning concepts and methods for addressing challenges in probabilistic hydrological post-processing and forecasting

Georgia Papacharalampous[1,*,†], and Hristos Tyralis[2,†]

[1] Department of Water Resources and Environmental Engineering, School of Civil Engineering, National Technical University of Athens, Heroon Polytechneiou 5, 15780 Zographou, Greece

[2] Construction Agency, Hellenic Air Force, Mesogion Avenue 227–231, 15561 Cholargos, Greece

\* Correspondence: papacharalampous.georgia@gmail.com, tel: +30 69474 98589

† These authors have contributed equally to this work and share first authorship

**Email addresses and ORCID profiles:** papacharalampous.georgia@gmail.com, gpapacharalampous@hydro.ntua.gr, https://orcid.org/0000-0001-5446-954X (Georgia Papacharalampous); montchrister@gmail.com, hristos@itia.ntua.gr, https://orcid.org/0000-0002-8932-4997 (Hristos Tyralis)

**Abstract:** Probabilistic forecasting is receiving growing attention nowadays in a variety of applied fields, including hydrology. Several machine learning concepts and methods are notably relevant towards addressing the major challenges of formalizing and optimizing probabilistic forecasting implementations, as well as the equally important challenge of identifying the most useful ones among these implementations. Nonetheless, practically-oriented reviews focusing on such concepts and methods, and on how these can be effectively exploited in the above-outlined essential endeavour, are currently missing from the probabilistic hydrological forecasting literature. This absence holds despite the pronounced intensification in the research efforts for benefitting from machine learning in this same literature. It also holds despite the substantial relevant progress that has recently emerged, especially in the field of probabilistic hydrological post-processing, which traditionally provides the hydrologists with probabilistic hydrological forecasting implementations. Herein, we aim to fill this specific gap. In our review, we emphasize key ideas and information that can lead to effective popularizations, as such an emphasis can support successful future implementations and further scientific developments. In the same forward-looking direction, we identify open research questions and propose ideas to be explored in the future.

## 1. Background information, basic terminology and review overview

"Prediction" is a broad and generic term that describes any process for obtaining guesses of unseen variables based on any available information, as well as each of these guesses. On the other hand, "forecasting" is a more specific term that describes any process for issuing predictions for future variables based on information (which most commonly takes the form of time series) about the present and the past, with these particular predictions being broadly called "forecasts". Forecasting is a key theme and topic for this study. Therefore, in what follows, the general focus will be on it and not on prediction in general, although many of the statements and methods that will be referring to it are equally relevant and applicable to other prediction types.

The origins of forecasting trace back to the early humans and their pronounced need for certainty in the practical endeavour of supporting their various everyday life decisions (Petropoulos et al. 2022). Thus, forecasting has met until today and still meets numerous implementations, formal and informal. Independently of their exact categorization and features, the formal implementations of forecasting rely, in principal, on concepts, theory and practice that originate from or can be attributed to the predictive branch of statistical modelling, although forecasting is also considered as an entire field on its own because of the major role that the temporal dependence plays in the formulation of its methods. The predictive branch of statistical modelling exhibits profound and fundamental differences with respect to the descriptive and explanatory ones, as it is thoroughly explained in Shmueli (2010). All these three statistical modelling branches and their various tasks are known to have utility and value themselves with no exceptions; still, the ultimate goal behind all of them, even behind (the majority of) the other prediction tasks, should be forecasting (Billheimer 2019).

Overall, the various forecasts can be categorized in many ways. Regular groupings of forecasts are those based on the forecast horizon or lead time. The most common relevant categories are the ones known under the terms "one-step ahead" and "multi-step ahead" forecasts (which appear extensively, for instance, in the general forecasting and the energy forecasting literatures; see, e.g., Bontempi and Taieb 2011; Taieb et al. 2012), as well as those known under the terms "real-time", "short-range", "medium-range" and "long-range" forecasts (which appear extensively, for instance, in the meteorological and hydrological forecasting literatures; see, e.g., Gneiting and Raftery 2005; Yuan et al. 2015). In the context of the same categorization rule, the terms "short-term", "medium-term" and "long-term" forecasts also appear broadly (see, e.g., Regonda et al. 2013; Yuan et al. 2015). Other meaningful groupings are based on the temporal scale at which the forecasting takes place. In this context, the various categories and terms range from the "sub-seasonal" to the "seasonal" or even the "annual" and "inter-annual" forecasts (see, e.g., Gneiting and Raftery 2005; Yuan et al. 2015). Obviously, the lead time and the temporal scale of the forecasts are related to each other. Another distinction between forecasts can be made based on whether they refer to continuous or categorical variables, with the former case consisting the most common one in the literature and, thus, also the general focus of this study.

In the context of another regular categorization rule, one category includes the best-guess forecasts, which are best guesses for future variables, each taking the form of a single value. These forecasts have been traditionally and predominantly supporting decision making in almost every applied field (Gneiting and Katzfuss 2014), including hydrology (Krzysztofowicz 2001). Their most common formal implementations for the case of continuous variables are the mean- (also known as "expected-") and the median-value forecasts, which are simply the mean and median values, respectively, of their corresponding predictive probability distributions (i.e., the probability distributions of the targeted future variables conditioned upon the data and models utilized for the forecasting; see, e.g., Gelman et al. 2013, for the mathematical formulation of this definition). Best-guess forecasts are else referred to in the forecasting literature as "best-estimate", "single-value", "single-valued", "single-point" or even more broadly as "point" forecasts, while sometimes they are additionally said to correspond to the "conditional expectation", the "conditional mean" or the "conditional median" of the future variable of interest (see, e.g., the terminology adopted for such forecasts in Holt 2004; Giacomini and Komunjer 2005; Gneiting 2011; Torossian et al. 2020; Hyndman and Athanasopoulos 2021, Chapter 1.7).

A best-guess forecast can be obtained by utilizing traditional and more modern time series (also referred to as "stochastic") models (e.g., those by Brown 1959; Winters 1960; Box and Jenkins 1970; Holt 2004; Hyndman and Khandakar 2008) or supervised machine and statistical learning algorithms for regression or classification (e.g., those listed and documented in Hastie et al. 2009; James et al. 2013) under proper formulations, which largely depend on the exact forecasting problem under consideration. Another well-established way for issuing best-guess forecasts in hydrological settings is based on the hydrological modelling literature and consists in running process-based rainfall-runoff models (i.e., models that are built based on process understanding for taking information on rainfall and other meteorological variables as their inputs to give runoff or streamflow in their output) in forecast mode (i.e., by using meteorological forecasts as inputs; Klemeš 1986). These models are also extensively exploited in simulation mode (i.e., by using meteorological observations as inputs; Klemeš 1986) and can be classified into conceptual and physically-based models (see, e.g., the relevant examples provided in Todini 2007, as well as the 36 conceptual rainfall-runoff models in Knoben 2020). Note here that the terms "simulation" and "prediction" are used as synonymous in the hydrological modelling literature (Beven and Young 2013). In what follows, we will be using the term "hydrological forecasting" to exclusively refer to the forecasting of runoff or streamflow variables (which, in its best-guess form, could be made, for instance, by following one of the above-outlined approaches) and their extreme behaviours, although the same term is also used relatively frequently in the literature for the forecasting of other hydrometeorological and hydroclimatic variables, such as the rainfall, water quality, soil moisture and water supply ones. The term "hydrological forecast" will be used accordingly.

The alternative to issuing best-guess forecasts is issuing probabilistic forecasts, which include almost always best-guess forecasts and, at the same time, provide additional information about the predictive probability distributions. More precisely, a probabilistic forecast can take either (i) the form of an entire predictive probability distribution ([Krzysztofowicz 2001](); [Gneiting and Katzfuss 2014]()), with Bayesian statistical models consisting the earliest formal procedures for obtaining this particular form (see, e.g., the work by [Roberts 1965]()), or (ii) comprehensively reduced forms that might include single quantile or interval forecasts (see, e.g., the remarks on the usefulness and importance of such forecasts in [Chatfield 1993](); [Giacomini and Komunjer 2005]()) or, more commonly, sets of such forecasts that might additionally comprise a mean-value forecast (see, e.g., the forecast examples in [Hyndman and Athanasopoulos 2021](), Chapter 1.7). Indeed, such reduced forms can effectively summarize the corresponding entire predictive probability distributions for technical applications. Simulations of the predictive probability distribution, which are usually obtained in Bayesian settings using Monte Carlo Markov Chain (MCMC)-based techniques, or characterizations of the predictive probability distribution using ensemble members can be said to belong to both the above categories ([Bröcker 2012]()).

A quantile forecast is simply a quantile of the corresponding predictive probability distribution and might also be referred to in the literature under the alternative terms "conditional quantile", "predictive quantile" or "forecast quantile" or even as a "point" forecast corresponding to a specific "quantile level" (see, e.g., the terminology adopted in [Giacomini and Komunjer 2005](); [Gneiting 2011]()). The latter level indicates the probability with which the quantile forecasts should exceed their corresponding actual future values. Moreover, an interval forecast is simply defined by two quantile forecasts, provided that these quantile forecasts correspond to different quantile levels, and is alternatively referred to under the terms "predictive interval" or "prediction interval" (see, e.g., the terminology adopted in [Chatfield 1993](); [Lichtendahl et al. 2013](); [Abdar et al. 221](); [Hyndman and Athanasopoulos 2021](), Chapter 1.7), with the most common prediction intervals being by far the central ones (i.e., intervals bounded by symmetric level quantiles). The $p$% (central) prediction intervals, with $p$ taking values larger than 0 and smaller than 100, are considered to have an optimal coverage (i.e., maximum reliability) if they include the actual future values with frequency $p$%. Examples of probabilistic hydrological forecasts are presented in [Figure 1]().

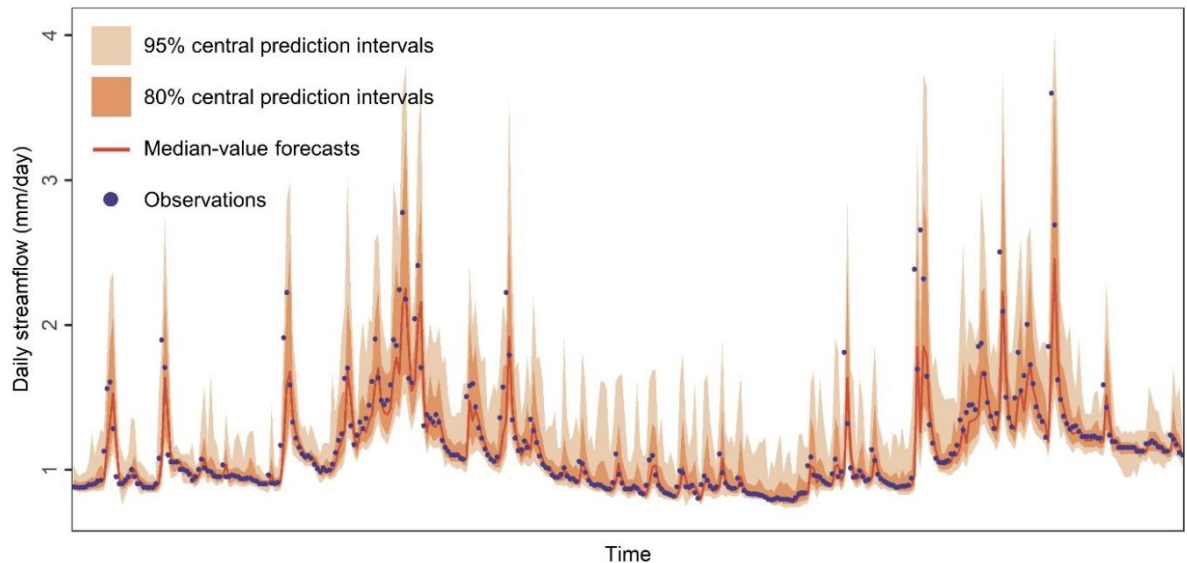

Figure 1. Probabilistic one-day ahead forecasts that are consisted of (i) median-value forecasts (depicted with a red line), (ii) 80% central prediction intervals (depicted with a dark orange ribbon) and (iii) 95% central prediction intervals (depicted with a light orange ribbon) for a daily streamflow time series (depicted with purple points).

Probabilistic forecasting in general, and probabilistic hydrological forecasting in particular, is receiving growing attention nowadays for multiple reasons that include: (a) the increasing embracement of the concept of predictive uncertainty (i.e., a fundamental mathematical concept that underlies probabilistic forecasting and has been thoroughly placed in a hydrological context by Todini 2007); and (b) the larger degree of information that the probabilistic forecasts can offer to the practitioners compared to the best-guess forecasts. Extensive discussions on this latter point can be found in Krzysztofowicz (2001). These discussions rotate around the rapidly expanding belief that probabilistic forecasts are an *"essential ingredient of optimal decision making"* (Gneiting and Katzfuss 2014), which also consists the key premise that underlies a variety of important research efforts and advancements, both in hydrology and beyond. In spite of these efforts and advancements, probabilistic forecasting is still a relatively new endeavour and, therefore, it carries with it numerous fresh challenges that need to be formally recognized and addressed in an optimal way (Gneiting and Katzfuss 2014). Perhaps the most fundamental, and at the same time universal across disciplines, umbrella methodological challenges from the entire pool are those of formalizing and optimizing probabilistic forecasting implementations, as well as that of identifying the most useful ones among the various implementations.

Machine learning concepts and methods can provide effective and straightforward solutions to these specific challenges. Indeed, we have recently witnessed the transfer of some notably useful machine learning concepts and methods in the field of probabilistic hydrological forecasting and, even more frequently, in its sister field of "probabilistic hydrological post-processing". This latter field can be defined as the one that comprises a wide range of statistical methods (which include but are not limited to machine learning ones) for issuing probabilistic hydrological forecasts and, more generally, probabilistic hydrological predictions by using the best-guess outputs of process-based rainfall-runoff models as predictors in regression settings (see, e.g., a review of such methods in Li et al. 2017). Although the term "post-processing" is sometimes also used to refer to best-guess procedures that are limited to improving the accuracy of best-guess outputs of process-based rainfall-runoff models, only its probabilistic version is relevant herein. This specific version relies on models that can offer accuracy improvements as well, but their utility in the overall framework is not limited to such improvements.

A considerable part of the "probabilistic hydrological post-processors" are (i) broadly referred to as methods for estimating (and advancing our perception of) the "uncertainty" of the various hydrological predictions or simulations (see, e.g., related definitions in Montanari 2011), and (ii) tested with the process-based rainfall-runoff model being run in simulation mode (see, e.g., the modelling works by Montanari and Brath 2004; Montanari and Grossi 2008; Solomatine and Shrestha 2009; Montanari and Koutsoyiannis 2012; Bourgin et al. 2015; Dogulu et al. 2015; Sikorska et al. 2015; Farmer and Vogel 2016; Bock et al. 2018; Papacharalampous et al. 2019; Tyralis et al. 2019a; Papacharalampous et al. 2020b; Sikorska-Senoner and Quilty 2021; Koutsoyiannis and Montanari 2022; Quilty et al. 2022; Li et al. 2021a; Romero-Cuellar et al. 2022). Notably, reviews, overviews and popularizations that focus on the above-referred to as existing and useful machine learning concepts and methods are currently missing from the probabilistic hydrological post-processing and forecasting literatures.

This scientific gap exists despite the large efforts being made towards summarizing and fostering the use of machine learning in hydrology (see, e.g., the reviews by Solomatine and Ostfeld 2008; Raghavendra and Deka 2014; Mehr et al. 2018; Shen 2018; Tyralis et al. 2019b) and in best-guess hydrological forecasting (see, e.g., the reviews by Maier et al. 2010; Sivakumar and Berndtsson 2010; Abrahart et al. 2012; Yaseen et al. 2015; Zhang et al. 2018). It also exists despite the comparably large efforts made for strengthening progress in the field of ensemble hydrological forecasting (see, e.g., the review by Yuan et al. 2015). This latter field (see, e.g., the methods in Regonda et al. 2013; Pechlivanidis et al. 2020; Girons Lopez et al. 2021; Liu et al. 2022) offers a well-established way of issuing operational probabilistic hydrological forecasts. In the related typical implementations, process-based rainfall-runoff models are utilized in forecast mode with ensembles of meteorological forecasts (which are created on a regular basis by atmospheric scientists to meet a wide range of applications; Gneiting and Raftery 2005) in their input to deliver an ensemble of point hydrological forecasts that collectively constitute the output probabilistic forecast.

In this work, we aim to fill the above-identified gap towards formalizing the exploitation of machine and statistical learning methods (and their various extensions) for probabilistic hydrological forecasting given hydrological and meteorological inputs that can be but are not necessarily the same to those required for ensemble hydrological forecasting. Indeed, only such a formalization can allow making the most of the multiple possibilities offered by the algorithmic modelling culture (see Breiman 2021b, for extensive and enlightening discussions on this culture and its implications) in practical probabilistic hydrological forecasting settings. For achieving our aim, we first summarize the qualities of a good probabilistic hydrological forecasting method in Section 2. We then select the most relevant machine learning concepts and methods towards ensuring these qualities, and briefly review their related literature in Section 3. In our review, we emphasize key ideas and information that can lead to effective popularizations and syntheses of the concepts and methods under investigation, as such an emphasis could support successful future implementations and further scientific developments in the field. In the same forward-looking direction, we identify open research questions and propose ideas to be explored in the future. Lastly, we summarize and conclude the work by further discussing its most important aspects in terms of practical implications in Section 4.

## 2. What is a good method for probabilistic hydrological forecasting

The title of this section emulates the successfully *"eye-catching"* title *"What is the 'best' method of forecasting?"* that was given to the seminal review paper by Chatfield (1988) from the forecasting field. This same paper begins by stating that the reader who expects a simple answer to the question consisting the paper's title might eventually get disappointed by the contents of the paper, although some general guidelines are still provided in it. Indeed, a universally best forecasting method does not exist and, therefore, instead of pursuing its proper definition and its finding, one should pursue making the most of multiple good forecasting methods by using, each time, the most relevant one (or ones) depending on exact formulation of the forecasting task to be undertaken (see, e.g., discussions by Jenkins 1982; Chatfield 1988). However, even the definition of a good forecasting method in terms of specific qualities can get quite challenging and is mostly equivalent to the definition of a useful forecasting method (see, e.g., discussions by Jenkins 1982; Chatfield 1988; Hyndman and Khandakar 2008; Taylor and Letham 2018), thereby rotating around a wide variety of considerations to be optimally weighed against each other in the important direction of effectively making the targeted future quantities and events more manageable on a regular basis for the practitioners.

Among these considerations, obtaining skillful probabilistic forecasts (with the term "skillful" and its relatives being used herein to imply high predictive performance from perspectives that do not necessarily involve skill scores; for the definition and examples of the latter, see Gneiting and Raftery 2007) is perhaps by far the easiest to perceive and recognize. In fact, probabilistic forecasting aims at reducing the uncertainty around predicted future quantities and events (with the importance of this target having been recognised in hydrology with the $20^{th}$ of the 23 major unsolved problems; Blöschl et al. 2019), similarly to what applies to best-guess forecasting, but from a conditional probabilistic standpoint. Thus, the more skillful the probabilistic forecasts, the less uncertain and the more manageable can become for the practitioner the predicted future quantities and events. Probabilistic predictions and forecasts are, in principle, considered to be skillful (again in a more general sense than the one relying on skill scores) when the sharpness of the predictive probability distributions is maximized, subject to reliability, on the basis of the available information set (Gneiting and Katzfuss 2014). This constitutes indeed the formal criterion for assessing probabilistic predictions and forecasts. In this criterion, the term "reliability" refers to the degree of coverage of the actual future values by the various prediction intervals (or the probability with which the quantile forecasts exceed their corresponding actual future values; see again related remarks in Section 1). Moreover, the term "sharpness" refers to how wide or narrow the predictive probability distributions and, thus, the various prediction intervals are. Scoring rules that are proper for the general task of probabilistic forecasting, with this propriety being evaluated strictly in terms of meeting the above criterion, include the quantile, interval and continuous ranked probability scoring rules, among others, with the latter of them being broadly referred to in the literature under its abbreviation "CRPS". These scoring rules and their documentations can be found, for instance, in Dunsmore (1968), Gneiting and Raftery (2007) and Gneiting (2011). Notably, scoring rules that evaluate either reliability or sharpness alone are not proper for the task; yet, they could have some usefulness in terms of interpreting proper comparative evaluations. Also notably, the same criterion is not appropriate for assessing the skill of the probabilistic forecasts of extreme events (Brehmer and Strokorb 2019), in a similar way that the root mean square error (RMSE) is not appropriate for assessing best-guess forecasts of extreme events (see also discussions in Tyralis and Papacharalampous 2022), and consequently the forecast evaluation in this special case necessarily reduces into the computation of scores that are not designed particularly for probabilistic forecasts (e.g., point summaries of the tails of the predictive probability distributions; Lerch et al. 2017) or it relies on the most recent developments for adjusting scoring rules to meet such special requirements (see, e.g., the developments by Taggart 2022).

Aside from skill, there are several additional, but still crucial, considerations driving the formulation and selection of forecasting methods that are mostly overlooked in the vast majority of research papers, both those appearing in hydrology and beyond. Indeed, not all the methodological developments can be exploited in technical and operational contexts, and even some of the most skillful probabilistic forecasting methods might not be useful in practice, at least considering the current limitations. Among the most characteristic considerations are, therefore, those for meeting the various requirements accompanying the ambitious, yet necessary and fully achievable, endeavour of forecasting "at scale" (Taylor and Letham 2018). These requirements have been enumerated and extensively discussed in the context of probabilistic hydrological post-processing and forecasting by Papacharalampous et al. (2019), and include those for (a) a massive number of forecasts and (b) a massive variety of "situations" and quantities to be forecasted, thereby imposing the development of fully automatic, widely applicable and computationally fast (or at least affordable) forecasting methods. Importantly, a large degree of automation should not be interpreted to imply a small degree of flexibility in the forecasting method's formulation, as the opposite should actually hold (see, e.g., the examples by Hyndman and Khandakar 2008; Taylor and Letham 2018). This form of flexibility is indeed required, for instance, in terms of dealing with diverse conditions of data availability (either for accelerating forecasting solutions by making the most of the available data, or even for assuring the delivery of forecasts in conditions of poor data availability). It should further ensure any proper adjustment and special treatment that might be required for achieving optimality in terms of skill and for dealing with special categories of future events and quantities (e.g., extremes).

Moreover, simplicity, straightforwardness, interpretability and explainability could also be recognized as qualities of a good forecasting method, although their definition is more subjective than the definition of other qualities (such as those of skill, applicability and automation) and their consideration (or not) largely depends on the forecaster and the user of the forecasts. In fact, these rather secondary and optional qualities could make the forecasts easier to trust and handle in practice, thereby making a forecasting method more useful (see, e.g., discussions by Chatfield 1988; Januschowski et al. 2020). Even further from such benefits, simplicity and straightforwardness could be additionally important in terms of minimizing the computational cost, while interpretability and explainability can also offer scientific insights, along with a solid ground for future methodological developments, and are considered particularly important in hydrology. Lastly, a probabilistic forecasting method is sometimes judged on the basis of the exact form of its outputs, specifically from whether these outputs take the form of entire predictive probability distributions or reduced forms (which, however, can resemble quite satisfactorily entire predictive probability distributions, provided that they comprise forecasts for multiple statistics of theirs), with the former form being somewhat easier to interpret and follow, especially for unexperienced users of the forecasts.

## 3. Machine learning for probabilistic hydrological forecasting

### 3.1 Quantile, expectile, distributional and other regression algorithms

There is a widespread misconception in hydrology that machine and statistical learning algorithms cannot provide probabilistic predictions and forecasts unless they get merged with other models within wider properly designed frameworks and, thus, a large amount of research efforts are devoted towards addressing this particular challenge. However, this challenge could be safely skipped (thereby saving research efforts for devoting them to other challenges) by adopting suitable developments that are originally made beyond hydrology, specifically those that are founded on the top of the pioneering concept of going "beyond mean regression" (Kneib 2013). Indeed, there are already whole families of machine and statistical learning regression algorithms that are explicitly designed to provide, in a straightforward and direct way, probabilistic predictions and forecasts. Also notably, a considerable portion of the implementations of these algorithms are made available in open source software after being optimally programmed by computer scientists and are equally user-friendly as the typical, broadly known regression algorithms (for mean regression). These families are outlined in the present section, with emphasis on the most well-received by the hydrological community and, at the same time, most practically appealing ones. Additional details on the similarities and differences between these same families with respect to their fundamentals and underlying assumptions can be found, for instance, in review paper by Kneib et al. (2021).

The quantile regression algorithms consist one of the most characteristic families for moving "beyond mean regression". These algorithms provide quantile predictions and forecasts (according to the definitions and illustrative examples provided in Section 1) in their output, and include, among others, the linear-in-parameters quantile regression algorithm (that is most commonly referred to simply as "quantile regression" in the literature; Koenker and Bassett 1978), as well as its autoregressive variant (Koenker and Xiao 2006) and additional extensions (see, e.g., their summary by Koenker 2017). Other typical examples of quantile regression algorithms (or, more generally, algorithms that can support quantile estimation, among other learning tasks) include the *k*-nearest neighbors for quantile estimation (Bhattacharya and Gangopadhyay 1990), quantile regression forests (Meinshausen 2006), generalized random forests for quantile estimation (Athey et al. 2019), gradient boosting machines (Friedman 2001), model-based boosting based on linear or non-linear models (Bühlmann and Hothorn 2007; Hofner et al. 2014) and quantile regression neural networks (originally introduced by Taylor 2000 and later improved by Cannon 2011), while there are also quantile autoregression neural networks (Xu et al. 2016), composite quantile regression neural networks (Xu et al. 2017), quantile deep neural networks (Tagasovska and Lopez-Paz 2019), composite quantile regression long short-term memory networks (Xie and Wen 2019), quantile regression long short-term memory networks with exponential smoothing components (Smyl 2020) and quantile regression neural networks for mixed sampling frequency data (Xu et al. 2021). Additional examples from this same algorithmic family include the XGBoost (eXtreme Gradient Boosting machine; Chen and Guestrin 2016) and LightGBM (Light Gradient Boosting Machine; Ke et al. 2017) algorithms for estimating predictive quantiles, the random gradient boosting algorithm (which combines random forests and boosting; Yuan 2015) and optimized versions from a practical point of view (Friedberg et al. 2020; Gasthaus et al. 2020; Moon et al. 2021).

As the above-reported names largely indicate, all these algorithms are close relatives (variants) of broadly known mean regression algorithms, such as the linear regression (see, e.g., documentations in James et al. 2013, Chapter 3), *k*-nearest neighbors (see e.g., documentations in Hastie et al. 2009, Chapter 2.3.2), random forests (Breiman 2001a), boosting algorithms (see e.g., documentations in Hastie et al. 2009, Chapter 10), neural networks (see e.g., documentations in Hastie et al. 2009, Chapter 11) and deep neural networks (Hochreiter and Schmidhuber 1997; Lecun et al. 2015). Therefore, similarly to them, their relative performance depends largely on the real-world problem that needs to be solved, and they can also differ notably with each other in terms of interpretability and flexibility (with these two algorithmic features being broadly recognized as incompatible to each other; see, e.g., James et al. 2013, Chapter 2.1.3, for characteristic examples on their trade-off). Indeed, they span from the most interpretable (least flexible) statistical learning ones (e.g., quantile regression) to less interpretable (more flexible) machine and deep learning ones (e.g., quantile regression forests and quantile deep neural networks). Theoretical details supporting their exact formulations can be found, for instance, in the references that are provided in the above paragraph or in the longer list of references in Torossian et al. (2020) and are out of the scope of this work, which is practically oriented.

Given this specific orientation, it is important to explain in simple terms the key idea behind the majority of the quantile regression algorithms. This specific idea was first conceived and successfully implemented by Koenker and Bassett (1978) for formulating the simplest quantile regression algorithm (Waldmann 2018) and is very simple itself. For its herein provided popularization, let us begin by supposing one of our most familiar problems, the typical (i.e., the mean) regression problem. For solving this specific problem, an algorithm needs to "learn" how the mean of the response variable changes with the changes of the predictor variables. For achieving this, the least-square error objective function or some other similarly conceptualized error function is routinely incorporated into the algorithm and guides its training by consisting the loss function that is minimized. Let us now suppose that we are not interested in the future mean of the response variable, but instead that we are interested in that future value of streamflow at time *t* that will be exceeded with probability 10%. In this case, the algorithm needs to "learn" how the streamflow quantile of level 0.90 changes with the changes of the predictor variables. For achieving this, the quantile scoring function (else referred to as "pinball loss" function in the literature; see, e.g., Gneiting and Raftery 2007; Gneiting 2011, for the its definition) can be incorporated (instead of the least-square loss function) into the algorithm for placing the focus on the targeted streamflow quantile, thereby effectively allowing the algorithm's straightforward training for probabilistic prediction and forecasting. This training is then made by exploiting the formal criterion for achieving skillful probabilistic predictions and forecasts (see Section 2).

In a nutshell, most of the quantile regression algorithms (e.g., the quantile regression, linear boosting, gradient boosting machine and quantile regression neural network ones) are trained by minimizing the quantile scoring function separately for each quantile level, while among the most characteristic examples of quantile regression algorithms that do not rely on this particular minimization, but on other optimization procedures, are the quantile regression forests and the generalized random forests for quantile regression. Independently of the exact optimization procedure applied, there exist clear guidance in the literature and, more precisely, in Waldmann (2018) on when quantile regression algorithms consist a befitting choice. In brief, this is the case when: (a) there is interest in events at the "limit of probability" (i.e., further than the most central parts of the predictive probability distributions); (b) there is no information at hand on which probability distribution models represent sufficiently the predictive probability distributions (or such information is hard to deduce); (c) there are numerous outliers among the available observations of the dependent variable; and (d) heteroscedasticity needs to be modelled. Based on the above-summarized guidance, we understand that probabilistic hydrological forecasting can indeed benefit from the family of quantile regression algorithms in the direction of issuing skillful probabilistic forecasts. In fact, several algorithms from this specific family have already been found relevant to this task and have been tested in the field of probabilistic hydrological post-processing, including the simplest linear-in-parameters (e.g., in Weerts et al. 2011; López López et al. 2014; Dogulu et al. 2015; Bogner et al. 2017; Wani et al. 2017; Papacharalampous et al. 2019; Tyralis et al. 2019a; Papacharalampous et al. 2020a,b) and several machine learning (e.g., in Bogner et al. 2016, 2017; Papacharalampous et al. 2019; Tyralis et al. 2019a) ones.

Probabilistic hydrological post-processing through quantile regression algorithms has been extensively discussed as a culture-integrating approach to probabilistic hydrological prediction and forecasting by Papacharalampous et al. (2019). The relevant discussions are primarily based on the overview by Todini (2007), in which the process-based rainfall-runoff models and the data-driven algorithms (with the latter including, among others, all the machine and statistical learning ones) are summarized as two different *"streams of thought"* (or cultures) that need to be harmonized *"for the sake of hydrology"*. A basic probabilistic hydrological post-processing methodology comprising a process-based rainfall-runoff mode and a quantile regression algorithm, is summarized in Figure 2. Notably, this methodology could be further enriched in multiple ways. For instance, information from multiple process-based rainfall-runoff models could be exploited (see, e.g., related discussions by Montanari and Koutsoyiannis 2012), while the same applies for other additional predictors. Such predictors could include various types of meteorological forecasts (and possibly entire ensembles of such forecasts), and past or present hydrological and meteorological observations. Of course, the utilization of best-guess hydrological forecasts as predictors (see again Figure 2) is not absolutely necessary, which practically means that probabilistic hydrological forecasting can be made without applying post-processing. Although both the absolute and relevant performance of probabilistic hydrological post-processors might (and is rather expected to) depend on whether the process-based rainfall-runoff model is applied in simulation or in forecast mode, the possible solutions and the various pathways towards addressing the challenges of formalizing, optimizing and selecting probabilistic hydrological post-processors using machine learning do not. Still, a clear distinction between these two modelling contexts is necessary when trying to benefit from past post-processing works for achieving optimal machine learning solutions. Similarly, the absolute and relevant performance of machine learning methods might depend on whether they are applied in post-processing or more direct probabilistic hydrological forecasting contexts.

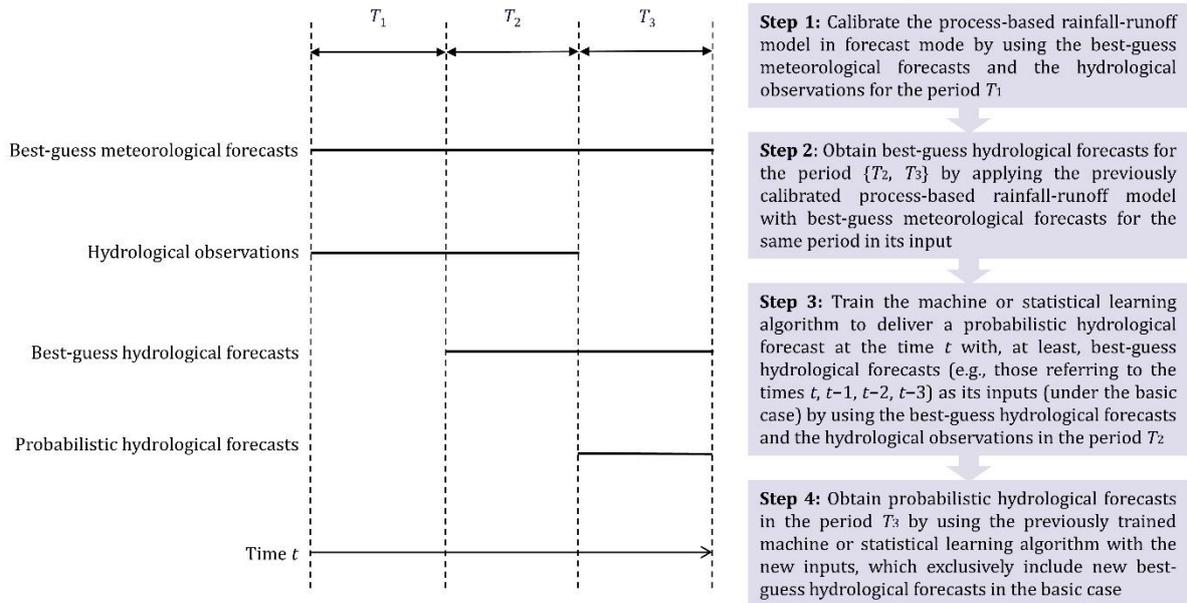

Figure 2. Basic formulation of probabilistic hydrological post-processing for providing probabilistic hydrological forecasts in the period $T_3$ using a process-based rainfall-runoff model in forecast mode (i.e., with meteorological forecasts as its inputs; Klemeš 1986) and a machine or statistical learning algorithm, along with best-guess meteorological forecasts and hydrological observations. The machine or statistical learning algorithm could originate from any of the families summarized in Section 3.1. For most of the quantile regression algorithms, steps 3 and 4 comprise in practice multiple trainings and runs, respectively, with each of them referring to a different quantile level, until all the quantile hydrological forecasts consisting the probabilistic hydrological forecast are issued. For example, for obtaining the probabilistic hydrological forecasts of Figure 1 through probabilistic hydrological post-processing using quantile regression neural networks (with the latter being the selected machine learning algorithm), steps 3 and 4 would comprise 5 trainings and 5 runs, respectively.

Overall, there are two ways for improving predictive performance: (i) improving the prediction algorithm; and (ii) discovering new informative predictors. For dealing with the latter of these challenges, the computation of variable importance scores (see, e.g., the reviews on the topic by Grömping 2015; Wei et al. 2015) is often suggested in the machine learning literature. Indeed, variable importance information helps us understand which predictors are the most relevant to improving predictive performance in each task (see, e.g., relevant investigations in a probabilistic hydrological post-processing context by Sikorska-Senoner and Quilty 2021), with this relevance being larger for the predictors with the larger scores. The variable importance scores can be classified into two main categories, which are known under the terms (Linardatos et al. 2020; Kuhn 2021): (a) "model-specific" (indicating that the application is restricted only to a specific family of algorithms); and (b) "model-agnostic" or "model-independent" (indicating that the application can be made in every possible algorithm). Their open source implementations are coupled with several machine and statistical learning algorithms (e.g., in linear models, random forests, partial least squares, recursive partitioning, bagged trees, boosted trees, multivariate adaptive regression splines – MARS, nearest shrunken centroids and cubist; Kuhn 2021). As perhaps proven by their popularity beyond hydrology, the most easy- and straightforward-to-compute for the task of probabilistic hydrological forecasting are those incorporated into the generalized random forest algorithm and into several boosting variants for quantile regression. Related summaries and literature information can be found in Tyralis et al. (2019b) and Tyralis and Papacharalampous (2021a). The various variable importance scores are implementations of the concept of explainable machine learning (see, e.g., the reviews on the topic by Linardatos et al. 2020; Roscher et al. 2020; Belle and Papantonis 2021), which is known for its utility for gaining scientific insights from machine learning algorithms, thereby achieving a deviation from "black box" solutions. Notably, this specific concept can take various additional forms of articulation and expression, which can also facilitate successful future implementations and an improved understanding of the forecasts by the practitioners.

In particular as regards the remarkably wide applicability characterizing the various quantile regression algorithms (most probably because of their non-parametric nature), with this particular applicability being sufficient even for meeting the strict operational requirements accompanying the endeavour of probabilistic hydrological forecasting, the reader is referred to the works by Papacharalampous et al. (2019) and Tyralis et al. (2019a). Indeed, these specific works present large-scale multi-site evaluations across the contiguous United States that could be possible only for widely applicable algorithms. The former of these works additionally presents the computational times spent for running the following six quantile regression algorithms in probabilistic hydrological post-processing: (i) quantile regression, (ii) generalized random forests for quantile regression, (iii) generalized random forests for quantile regression emulating quantile regression forests, (iv) gradient boosting machine based on trees, (v) boosting based on linear models and (vi) quantile regression neural networks. It further provides detailed information on how to implement these algorithms using open source software.

Aside from the above-discussed advantages, quantile regression algorithms also share a few characteristic drawbacks (Waldmann 2018). Indeed, the vast majority of these algorithms estimate separately predictive quantiles at different quantile levels. This implies some inconvenience in their utilization (in the sense that additional automation is required with respect to that already incorporated in the various open software implementations). Most importantly, it can cause quantile crossing, which however can be treated ad hoc by the forecaster (with this treatment requiring additional automated procedures). Also notably, parameter estimation is harder in quantile regression than in standard regression, while another drawback of quantile regression algorithms that is worthy to discuss is their inappropriateness for dealing with the important challenge of predicting extreme quantiles. Indeed, this inappropriateness could be a crucial limitation for the case of flood forecasting. Still, in such cases, proper extensions and close relatives of quantile regression algorithms could be applied. Indeed, Tyralis and Papacharalampous (2022) have recently proposed a new probabilistic hydrological post-processing method based on extremal quantile regression (Wang et al. 2012; Wang and Li 2013). This method extrapolates predictions issued by the quantile regression algorithm to high quantiles by exploiting properties of the Hill's estimator from the extreme value theory, while similar extensions for other quantile regression algorithms could also be possible. Another worth-mentioned modification that can be applied to any of these algorithms, inspired by the already existing quantile regression long short-term memory networks with exponential smoothing components by Smyl (2020), is the addition of a trend component for dealing with changing weather conditions beyond variability. In fact, the latter can be modelled directly by the quantile regression algorithms.

Other close relatives of the quantile regression algorithms are the expectile regression ones, which focus on conditional expectiles instead of conditional quantiles. Expectiles are the least squares analogues of the quantiles. Indeed, they are generalizations of the mean, in the same way that the quantiles are generalizations of the median. Among the existing expectile regression algorithms are the expectile regression (Newey and Powell 1987) and the expectile regression neural networks (Jiang et al. 2017). Notably, expectile regression algorithms (in their original forms) are a completely unexplored topic for probabilistic hydrological post-processing and forecasting contexts. Therefore, future research could investigate their place and usefulness in such contexts. Overall, it might be important to note that, similar to what applies for the quantile regression algorithms, the expectile regression algorithms should be expected to be more useful in the cases where there is no sufficient information at hand about the predictive probability distributions.

On the contrary, however, in the cases where such information exists, the distributional (else known as "parametric") algorithms should be expected to excel. These algorithms include several machine and statistical learning ones, which are usually referred to under the term "distributional regression" algorithms. Among them are the GAMLSS (Generalized Additive Models for Location, Scale and Shape; Rigby et al. 2005) and its extension in Bayesian settings, i.e., the BAMLSS (Bayesian Additive Models for Location, Scale, and Shape; Umlauf et al. 2018), as well as the distributional regression forests (Schlosser et al. 2019), a boosting GAMLSS model (Mayr et al. 2012), the NGBoost (Natural Gradient Boosting) model for probabilistic prediction (Duan et al. 2020) and the Gaussian processes for distributional regression (Song et al. 2019). Other distributional regression algorithms are the deep distribution regression algorithm (Li et al. 2021b), the marginally calibrated deep distributional regression algorithm (Klein et al. 2021) and the DeepAR model for probabilistic forecasting with autoregressive recurrent networks (Salinas et al. 2020). Notably, distributional regression algorithms could be also modified similarly to the quantile regression ones for dealing in an improved way with the special case of changing conditions beyond variability. Moreover, they could be applied with heavy-tailed distributions for approaching the other special case of predicting the extreme quantiles of the real-world distributions (that appear due to weather and climate extremes), thereby consisting alternatives to the previously discussed extensions of quantile regression based on the extreme value theory (see again Tyralis and Papacharalampous 2022).

Aside from the machine and statistical learning algorithms belonging to the above-outlined families, there are numerous others that can also provide probabilistic predictions and forecasts in a straightforward way (at least for dealing with the casual forecasting cases, while their modification or extension based on traditional statistics may be possible for special cases, such as changes beyond variability and extremes). Such algorithms are the BART (Bayesian Additive Regression Trees; Chipman et al. 2012) and their heteroscedastic variant (Pratola et al. 2020), which are regarded as boosting variants. Another machine learning algorithm that is notably relevant to the task of probabilistic hydrological forecasting is the dropout ensemble (Srivastava et al. 2014).

This deep learning algorithm has been proved to be equivalent to Bayesian approximation (Gal and Ghahramani 2016) and has already been proposed for estimating predictive hydrological uncertainty by Althoff et al. (2021). Its automated variant for probabilistic forecasting can be found in Serpell et al. (2019), while open software implementations of many other models, mostly deep learning ones, can be found in Alexandrov et al. (2020). Also notably, comprehensive reviews on deep learning and neural networks for probabilistic modelling can be found in Lampinen and Vehtari (2001), Khosravi et al. (2011), Abdar et al. (2021) and Hewamalage et al. (2021), where the interested reader can find numerous new candidates for performing probabilistic hydrological forecasting, thereby enriching the deep learning toolbox whose compilation has just started in the field (see, e.g., relevant works by Althoff et al. 2021; Li et al. 2021a).

Lastly, it is important to highlight that benefitting from the field of machine learning does not only include the identification and transfer (and perhaps even the modification) of various relevant machine and statistical learning algorithms. Indeed, more abstract inspirations sourced from this field can also lead to useful practical solutions. Characteristic examples of such inspirations are the concepts of "quantile-based hydrological modelling" (Tyralis and Papacharalampous 2021b) and "expectile-based hydrological modelling" (Tyralis et al. 2022). These concepts offer the most direct and straightforward probabilistic hydrological forecasting solutions using process-based rainfall-runoff models. In fact, the latter can be simply calibrated using the quantile or the expectile loss function for delivering quantile or expectile hydrological predictions and forecasts.

## 3.2   The "no free lunch" theorem on the a priori distinction between algorithms

In practice, several themes are routinely integrated for the formulation and selection of skillful forecasting methods. Among them are those for exploiting sufficient information that we might already have at hand about the exact forecasting problem to be undertaken and the various methods composing our toolbox. Characteristic examples of such themes appear extensively in previous sections and in the literature, and include the a priori distinction and selection of models based on their known properties, as well as the inclusion of useful data for the present and the past (and perhaps also the inclusion of useful forecasts) in the input of the various models. Indeed, in the above section we referred to the guidance by Waldmann (2018) for summarizing when the various quantile regression algorithms should be viewed as befitting modelling choices and when they should be expected to be more skillful than distributional regression or other distributional methods for probabilistic prediction and forecasting (e.g., the Bayesian ones in Geweke and Whiteman 2006), and vice-versa. We also referred to the concept of explainable machine learning and its pronounced relevance to the well-recognized endeavour of discovering new informative predictor variables for our algorithms. Even further, we highlighted the fact that quantile regression algorithms do not consist optimal choices (in their original forms) when extreme events need to be predicted and forecasted, and discussed how the capabilities of these same algorithms can be extended into the desired direction.

Themes such as the above are undoubtedly important in the endeavour of formulating and selecting skillful forecasting methods. Yet, it is equally important for the forecaster to recognize the following fact: Such themes and tools can only guide us through parts of the entire way and this is probably why a different family of themes is also routinely exploited towards an optimal selection between forecasting models. This latter family is founded on the top of scoring rules, and includes themes such as those of "forecast evaluation", "empirical evaluation", "empirical comparison" and "benchmarking". Indeed, the a priori distinction between models based on theoretical properties is not possible most of the times and, even when it is, it cannot always optimally support the selection between models. In fact, our knowledge on which properties matter the most in achieving optimal practical solutions might be limited or might be based on hardly relevant assumptions that should be avoided (e.g., assumptions stemming from descriptive or explanatory investigations, while the focus should be in the actual forecasting performance; see relevant discussions in Shmueli 2010). On the top of everything else, the various model properties are analytically derived and hold for infinite samples, while the samples used as inputs for forecasting are finite. Based on the above considerations, empirical evaluations, comparisons and benchmarking of forecasting models cannot be skipped when we are interested at maximizing the skill.

Importantly, there is a theorem underlying the discussions of this section, which is known as the "no free lunch" theorem (Wolpert 1996). This theorem is of fundamental importance for conducting proper benchmark evaluations and comparisons of methods for forecasting (and not only), and it was originally formulated for machine and statistical learning algorithms, as there are indeed whole groups of such algorithms that cannot be distinguished with each other to any extent, regarding their skill, based on their theoretical properties. In simple terms, the "no free lunch" theorem implies that, among the entire pool of relevant algorithms for dealing with a specific problem type (which, for the case of probabilistic hydrological forecasting, might include the various quantile regression algorithms that are enumerated in Section 3.1), there is no way for someone to tell in advance with certainty which of them will perform the best for one particular problem case (e.g., within a specific probabilistic hydrological forecasting case study). Indeed, there is "no free lunch" in the utilization of any machine or statistical learning algorithm, as there is "no free lunch" in the utilization of any model. Notably, the "no free lunch" theorem also implies that any empirical evidence that we might have for a specific case study cannot be interpreted as general empirical evidence and, therefore, forecast comparisons within case studies cannot support optimal selections between models, as it is also thoroughly explained in Boulesteix et al. (2018); nonetheless, there are still ways for the forecasters to deal with the "no free lunch" theorem in a meaningful sense. The most characteristic of these ways are discussed in detail in Sections 3.3 and 3.4.

### 3.3 Massive multi-site datasets and large-scale benchmarking

An effective way for dealing with the "no free lunch" theorem towards maximizing the benefits and reducing the risks, in terms of predictive skill, of machine and statistical learning algorithms (and not only) is through the concept of "large-scale benchmarking". This concept relies on the use of massive datasets (i.e., datasets that comprise multiple and diverse real-world cases, and sometimes also simulated toy cases) and multiple models, with the latter necessarily including benchmarks (e.g., simple or traditional models). Large-scale benchmarking consists the main concept underlying all the "large-scale comparison" studies, which are incomparably fewer than the "model development" studies in all the disciplines, while the opposite should actually hold to ensure that the strengths and limitations of the various models are well-understood and well-handled in practice (Boulesteix et al. 2018). It is also the core concept of the "M", "Kaggle" and other well-established series of competitions that appear in the forecasting and machine learning fields. Such competitions have a vital utility towards providing the community with properly formulated, independent and, therefore, also highly trustable evaluations of widely applicable and fully automated forecasting and/or machine learning methods. They are extensively discussed (by, e.g., Fildes and Lusk 1984; Chatfield 1988; Clements and Hendry 1999; Armstrong 2001; Fildes and Ord 2002; Fildes 2020; Athanasopoulos and Hyndman 2011; Castle et al. 2021; Januschowski et al. 2021; Lim and Zohren 2021; Makridakis et al. 2021) and reviewed (by, e.g., Hyndman 2020; Bojer and Meldgaard 2021) beyond hydrology, where the interested reader can find details about their history and characteristics. In particular as regards the fundamental necessity of utilizing benchmarks in forecast evaluation works, the reason behind it can be found in the outcomes of the already completed competitions. Indeed, simple (or less sophisticated) methods might perform surprisingly well in comparison to more sophisticated methods in some types of forecasting (and other) problems (Hyndman and Athanasopoulos 2021, Chapter 5.2).

In what follows, discussions on the practical meaning of large-scale benchmarking are provided. For these discussions, let us suppose a pool of probabilistic prediction methods from which we wish to select one (or more) for performing probabilistic hydrological forecasting (e.g., through post-processing). Among others, these candidate methods could include multiple machine and statistical learning ones, with each being defined not only by a specific algorithm (e.g., one of those enumerated in Section 3.1) but also by a specific set of predictor variables and by specific parameters (which are also commonly referred to as "hyperparameters"), or alternatively by the algorithm and automated procedures for predictor variable and parameter selection. For achieving optimal practical solutions in this particular context, we specifically wish to know the probabilistic hydrological forecasting "situations" in which it is more likely for each candidate to work better than the remaining ones. Note here that the various probabilistic hydrological forecasting "situations" of our interest could be defined by all the time scales and forecast horizons of our interest, by all the situations of data availability with which we might have to deal in practice, by all the quantile and prediction interval levels of our interest, by all the streamflow magnitudes of our interest or by other hydroclimatic conditions (e.g., those defined as "climate zones" by the various climate classification systems), or even by all these factors and several others. Since there is no theoretical solution to the above-outlined problem (see again discussions in Section 3.2), we can only provide empirical solutions to it. These latter solutions can, then, be derived through extensively comparing the performance of all the candidates in a large number and wide range of problem cases, which should collectively well-represent the various types of probabilistic hydrological forecasting "situations" being of interest to us. That is what large-scale benchmarking is, in its full potential, and that is why its value should be apprised in the direction of making the most of multiple good methods (e.g., for probabilistic hydrological forecasting) and not in the direction of selecting a single "best" method (see again Section 2).

In brief, if we empirically prove through large-scale benchmarking that a probabilistic hydrological forecasting method performs on average better in terms of skill than the remaining ones (see again Section 2, for a summary of the criteria and scoring rules that should guide such comparisons) for a sufficiently large number of cases representing a specific type of probabilistic hydrological forecasting "situations", then we have found that it is "safer" to use this specific method than using any of the remaining ones for this same type of probabilistic hydrological forecasting "situations" in the future. By repeating this procedure for all the possible categories of probabilistic hydrological forecasting "situations" (after properly defining them based on the various practical needs), the forecaster can increase the benefits and reduce the risks stemming from the use of multiple probabilistic hydrological forecasting methods. Given this pronounced relevance of large-scale benchmarking towards maximizing predictive skill, ensuring compliance with the various practical considerations accompanying the endeavour of formulating and selecting probabilistic hydrological forecasting methods (see again Section 2) gains some additional importance. Indeed, only the widely applicable, fully automated and computationally fast methods can be extensively investigated and further improved, if necessary, before applied in practice (or perhaps even discarded, but still having served a purpose as benchmarks for others). These same methods are also the only whose long-run future performance can be known in advance to a large extent, and include many machine and statistical learning ones, a considerable portion of which are enumerated in Section 3.1.

At this point, it is also important to highlight that there are multiple open multi-site datasets comprising both hydrological and meteorological information, thereby being appropriate for performing large-scale benchmarking (at least for the daily and coarser temporal scales) in probabilistic hydrological post-processing and forecasting (e.g., those by Newman et al. 2015; Addor et al. 2017; Alvarez-Garreton et al. 2018; Chagas et al. 2020; Coxon et al. 2020; Fowler et al. 2021; Klingler et al. 2021). Examples of studies utilizing such datasets to support a successful formulation and selection of skillful probabilistic hydrological forecasting or, more generally, probabilistic hydrological prediction methods also exist. Nonetheless, such examples mostly refer to single-method evaluations (or benchmarking) either in post-processing contexts (e.g., those in Farmer and Vogel 2016; Bock et al. 2018; Papacharalampous et al. 2020b) or in ensemble hydrological forecasting contexts (e.g., those in Pechlivanidis et al. 2020; Girons Lopez et al. 2021).

Ensemble hydrological forecasting can be (mostly) regarded as the well-established alternative in operational hydrology to the machine learning methods outlined in Section 3.1, independently of whether or not some type of post-processing is involved in the overall framework for probabilistic forecasting (or prediction). Still, some common data-related challenges are shared between these alternatives, as machine learning, too, should ideally be informed by "state-of-the-art" datasets comprising weather and/or climate forecasts to be then applied in operational mode. Although there are massive datasets comprising meteorological and hydrological forecasts and observations (see, e.g., those investigated in Pechlivanidis et al. 2020; Girons Lopez et al. 2021), the methods for ensemble weather forecasting keep updating. In such cases, the data that are actually available for (i) the training of the machine learning algorithms and (ii) the calibration of the hydrological models might be changing over time. Dealing with this particularity is somewhat more critical for forecasting with machine learning algorithms, because of the well-known importance of large data samples for their training. Approaches referred to under the term "online learning" (see, e.g., Martindale et al. 2020) could partly serve towards this important direction and could, thus, be investigated in this endeavour. Such approaches do not require a static dataset.

As regards the examples of large-scale comparisons and benchmarking of multiple machine and statistical learning methods, these are even rarer in the field, with the ones conducted in probabilistic hydrological post-processing contexts being available in Papacharalampous et al. (2019) and Tyralis et al (2019a). These two works can effectively guide the application of quantile regression algorithms in probabilistic hydrological forecasting. Indeed, although their large-scale results refer exclusively to the modelling "situations" determined by their experimental settings (i.e., the daily temporal scale, a specific process-based rainfall-runoff model, specific conditions of data availability and predictors, etc.), the re-formulation and extension of their methodological contribution to other hydrological forecasting settings would be a straightforward process, from an algorithmic point of view, and could be made in the future to answer those research questions that are still open on the relative performance of the various quantile regression algorithms in the field.

Of course, many more open research questions exist and concern the various families of machine learning algorithms that are discussed in Section 3.1, with some of them being completely unexplored. In particular, it would be useful for the hydrological forecaster to know how these families and their algorithms compare with each other, as well as to other families and their methods, with the latter possibly including several well-established alternatives that do not originate from the machine learning literature (e.g., the traditional Hydrologic Model Output Statistics – HMOS method; Regonda et al. 2013). Indeed, such information is currently missing from the literature. The various comparisons could be conducted, both in terms of skill and in more practical terms, in probabilistic hydrological forecasting for the various modelling "situations" exhibiting practical relevance, as this would allow making the most of the entire available toolbox in technical and operational settings. For achieving this, it would also be useful to deliver large-scale findings on predictor variable importance through explainable machine learning (see again the related remarks in Section 3.1), as such results could replace automated procedures for predictor variable selection with (mostly) satisfactorily results, thereby saving considerable time in operational settings. Massive multi-site datasets could also support hyperparameter selection investigations. Although these latter investigations could, indeed, be beneficial, it might be preferable to skip them (in favour of addressing other research questions) by simply selecting the predefined hyperparameter values that have been made available in the various open source implementations of the algorithms. According to Arcuri and Fraser (2013), these specific values are expected to lead to satisfactory performance in most cases (probably because they are selected based on extensive experimentation).

## 3.4 Forecast combinations, ensemble learning and meta-learning

Another way for dealing with the "no free lunch" theorem in a meaningful sense is based on the concept of "ensemble learning", a pioneering concept appearing in the community of machine learning that is equivalent to the concept of "forecast combinations" from the forecasting field. In forecasting through ensemble learning, instead of using one individual algorithm, an ensemble of algorithms is used (see, e.g., the seminal paper by Bates and Granger 1969, and the review papers by Clemen 1989; Granger 1989; Timmermann 2006; Wallis 2011; Sagi and Rokach 2018; Wang et al. 2022). The latter algorithms are known as "base learners" in the machine learning field, and are trained and then applied in forecast mode independently of each other. Their independent forecasts are finally combined with another learner, which is known as the "combiner" and is "stacked" on top of the base learners, with the final output being a single forecast (and the independent forecasts provided by the base learners being discarded after their combination). Notably, the term "ensemble learning" should not be confused with the terms "ensemble simulation" and "ensemble forecast" (or the term "ensemble forecasting"), which refer to formulations in which the entire ensemble of independent simulations or forecasts constitutes the probabilistic prediction or forecast (see, e.g., the model output forms in Montanari and Koutsoyiannis 2012; Hemri et al. 2013; Sikorska et al. 2015; Quilty et al. 2019; Pechlivanidis et al. 2020; Girons Lopez et al. 2021).

The simplest form of ensemble learning and "stacking" of algorithms is simple averaging, in which the combiner does not have to be trained, as it simply computes the average of the forecasts of the various base learners. For instance, the forecasts of quantile regression, quantile regression forests and quantile regression neural networks for the streamflow quantile of level 0.90 (i.e., three forecasts) can be averaged to produce a new forecast (i.e., one forecast), while the averaging of distributions is also possible. Some quite appealing properties and concepts are known to be related to simple averaging. Among them are the "wisdom of the crowd" and the "forecast combination puzzle". The "wisdom of the crowd" can be harnessed through simple averaging (Lichtendahl et al. 2013; Winkler et al. 2019) to increase robustness in probabilistic hydrological post-processing and forecasting using quantile regression algorithms (see related empirical proofs in Papacharalampous et al. 2020b) or potentially machine and statistical learning algorithms from the remaining families that are summarized in Section 3.1. By increasing robustness, one reduces the risk of delivering poor quality forecasts at every single forecast attempt. Overall, simple averaging is known to be hard to beat in practice, in the long run, for many types of predictive modelling "situations" (see relevant discussions by De Gooijer and Hyndman 2006; Smith and Wallis 2009; Lichtendahl et al. 2013; Graefe et al. 2014; Hsiao and Wan 2014; Winkler 2015; Claeskens et al. 2016), thereby leading to the challenging puzzle of beating this simple form of stacking with more sophisticated stacking (Wolpert 1992) and meta-learning forecast combination methods. Alternative possibilities for combining forecasts include Bayesian model averaging (see, e.g., Hoeting et al. 1999, for a related historical and tutorial review); yet, stacking has been theoretically proved to have some optimal properties in comparison to this alternative when the focus is on predictive performance (Yao et al. 2018).

In hydrology, the concept of ensemble learning has been extensively implemented for combining both best-guess forecasts (by, e.g., Diks and Vrugt 2010; Xu et al. 2018; Huang et al. 2019; Papacharalampous and Tyralis 2020; Tyralis et al. 2021) and probabilistic predictions (by, e.g., Vrugt and Robinson 2007; Hemri et al. 2013; Bogner et al. 2017; Papacharalampous et al. 2019; Tyralis et al. 2019a; Papacharalampous et al. 2020a,b; Li et al. 2022), with the Bayesian model averaging implementation being by far the most popular one. Probabilistic hydrological predictions of different machine learning quantile regression algorithms have been combined through simple averaging (by Papacharalampous et al. 2019; Tyralis et al. 2019a) and through stacking (by Tyralis et al. 2019a) in the context of probabilistic hydrological post-processing, and related large-scale benchmark tests have also been performed (by the same works). These benchmark tests stand as empirical proofs that simple averaging and stacking can offer considerable improvements in terms of skill in probabilistic hydrological prediction at the daily time scale, while similar large-scale investigations for the most specific case of probabilistic hydrological forecasting at the same and at other time scales with large practical relevance (and for various conditions of data availabilities) could be the subject of future research. Such investigations could also focus on the combination of probabilistic hydrological forecasts that have been previously issued based on different members of ensemble meteorological forecasts. Of course, the overall benefit from the use of ensemble learning methods should be also apprised again according to Section 2. Aside from the secondary considerations enumerated in this latter section, which are indeed met to a considerably lesser degree when forecast combinations are performed, the remaining considerations can be met quite satisfactorily, yet to a degree that largely depends on the choice of the base learners. That additionally implies that, ideally, the various combiners should be tested with as many different sets of base learners as possible in the context of large-scale benchmarking for optimizing long-run forecasting skill (see, e.g., the experimental setting in Papacharalampous and Tyralis 2020). Also notably, large-scale benchmark tests that examine the combination of entire predictive probability distributions are still missing from the hydrological literature and are, thus, recommended as future research.

Lastly, discussions should focus on the meta-learning approach to forecasting (see, e.g., some of the first relevant formulations for performing best-guess forecasting by Wang et al. 2009; Lemke and Gabrys 2010; Matijaš et al. 2013; Montero-Manso et al. 2020; Talagala et al. 2021). This approach is built on the reasonable premise that improvements in terms of skill can be obtained by conditioning upon time series features the weights with which the forecast combination is performed. This relatively recent idea can be interpreted in the sense that one method might be more skilful than others in forecasting time series with specific ranges of characteristics (with these characteristics standing as a new additional way for defining various modelling "situations" of interest for the forecasters) and implies the automation of practical forecasting systems that are necessarily trained through large-scale benchmarking in the direction of making the most of multiple forecasting methods (see again Section 3.3). Among the most typical time series features are the various autocorrelation, partial autocorrelation, long-range dependence, entropy, temporal variation, seasonality, trend, lumpiness, stability, nonlinearity, linearity, spikiness and curvature features (see the numerous examples in Wang et al. 2006; Fulcher et al. 2013; Fulcher and Jones 2014; Hyndman et al. 2015; Kang et al. 2017; Fulcher 2018; Kang et al. 2020), while the length and time scale of a time series could also be viewed as its features. Such general-purpose time series features for data science have been found relevant in interpreting the skill of best-guess hydrological forecasts at the monthly temporal scale in Papacharalampous et al. (2022), and are of fundamental and practical interest in hydrology (see, e.g., the central themes, concepts and directions provided by Montanari et al. 2013), especially in its stochastic branch.

Still, meta-learning consists a completely unexplored endeavour for the sister fields of probabilistic hydrological post-processing and forecasting. Given that the benefits from it could be considerably large (see again previous successful formulations for best-guess forecasting in Wang et al. 2009; Lemke and Gabrys 2010; Matijaš et al. 2013; Montero-Manso et al. 2020; Talagala et al. 2021), future research could be devoted to its exploration at the various temporal scales exhibiting practical relevance and for various data availability conditions. For this particular exploration, a variety of probabilistic forecasting methods (including, among others, those relying on the algorithms mentioned in Section 3.1) and a variety of time series features could be considered. It is, lastly, highly relevant to note that meta-learning methods for probabilistic hydrological forecasting could also be formulated around hydrological signatures, which have already been used for interpreting, from a process-oriented perspective, the performance of probabilistic hydrological forecasting methods by Pechlivanidis et al. (2020) and Girons Lopez et al.

(2021). Hydrological signatures are, indeed, the analogous of time series features in the catchment hydrology field, where the interested reader can find details about them (see, e.g., their taxonomies in McMillan et al. 2017 and McMillan 2020).

## 4. Summary, discussion and conclusions

Machine learning can provide straightforward and effective methodological solutions to many practical problems, including various probabilistic prediction and forecasting ones. With this practically-oriented review, we believe to have enriched the hydrological forecaster's toolbox with the most relevant machine learning concepts and methods for addressing the following major challenges in probabilistic hydrological forecasting: (a) how to formalize and optimize probabilistic forecasting implementations; and (b) how to identify the most useful among these implementations. The machine learning concepts and methods are summarized in Figure 3. We have thoroughly reviewed their literature by emphasizing key information that can lead to effective popularizations. We have also assessed the degree to which the field has already benefitted from them, and proposed ideas and pathways that could bring further scientific developments by also building upon existing knowledge, traditions and practices. The proposed pathways include both formal (and, thus, quite strict) ones and more abstract inspirations sourced from the machine learning field. Most importantly, we have proposed a united view that aims at making the most of multiple (typically as many as possible) methods, including but not limited to machine learning ones, by maximizing the benefits and reducing the risks from their use.

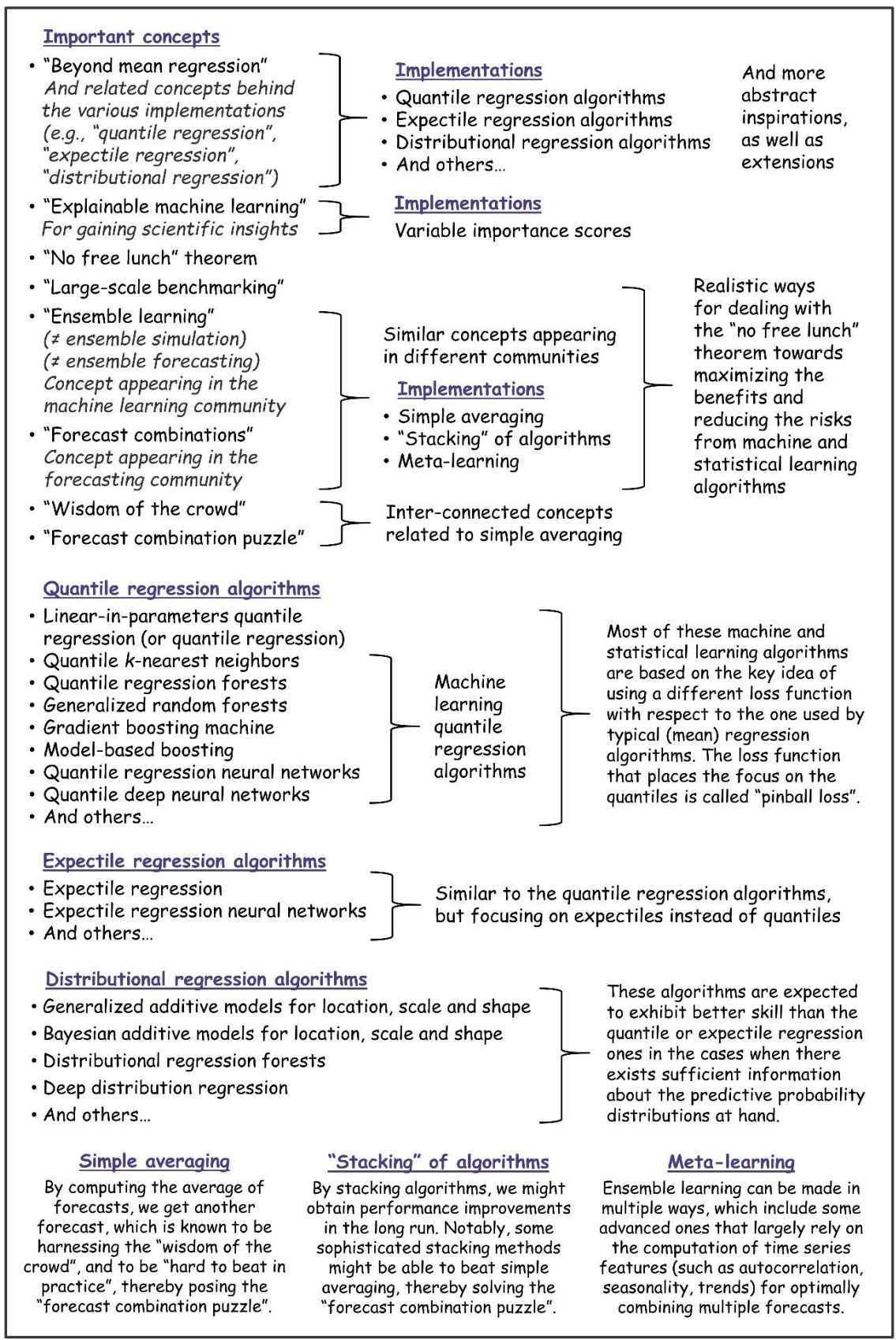

Figure 3. A practically-oriented synthesis of the most relevant machine learning concepts and methods for performing probabilistic hydrological post-processing and forecasting. The various procedures consisting probabilistic hydrological post-processing, in its basic formulation, are summarized in Figure 2.

Fostering research efforts under this united view is indeed particularly important, especially when the aim is at maximizing predictive skill. Natural companions in such a demanding endeavour are open science and open research (see, e.g., the related guidance in Hall et al. 2022), which are often overlooked in practice despite their vital significance. This review extensively discussed, among others, the fundamental relevance of massive open datasets to identifying the strengths and limitations of the various methods (both of the already available and the newly proposed ones) towards accelerating probabilistic hydrological forecasting solutions based on key discussions by Boulesteix et al. (2018), and based on the concept behind the forecasting and machine learning competitions. It also highlighted the importance of open software (see, e.g., packages in the `R` and `Python` programming languages, which are documented in R Core Team 2022 and Python Software Foundation 2022, respectively) for enriching the toolbox of the hydrological forecaster with algorithms that are optimally programmed (in many cases by computer scientists) and widely tested in various modelling "situations" before released. Overall, we believe that the summaries of the guidelines and considerations provided by this review are equally (and perhaps even more) important than the summaries of the various algorithms that are also provided. We would, therefore, like to conclude by emphasizing the need for formalizing research efforts as these guidelines and considerations imply.

**Author contributions:** The two authors have contributed equally to all the aspects of this work. They have made a substantial, direct, and intellectual contribution to the work and approved it for publication.

**Acknowledgements:** The authors are sincerely grateful to the Research Topic Editors for inviting the submission of this paper, to the Handling Editor for his additional work on it, and to the Reviewers for their constructive suggestions and remarks. Portions of this paper have been discussed by the authors in a popular science fashion in the HEPEX (Hydrologic Ensemble Prediction EXperiment) blog post entitled "Machine learning for probabilistic hydrological forecasting". This blog post is available online at the following link: https://hepex.inrae.fr/machine-learning-for-probabilistic-hydrological-forecasting.